\documentclass[runningheads]{llncs}

\usepackage{eccv}

\usepackage{bbding}

\usepackage{eccvabbrv}

\usepackage{graphicx}
\usepackage{booktabs}

\usepackage{algorithm}
\usepackage{algorithmic}
\usepackage{verbatim}

\newcommand*\samethanks[1][\value{footnote}]{\footnotemark[#1]}
\usepackage[accsupp]{axessibility}  

\usepackage{multirow}
\usepackage{hyperref}

\usepackage{orcidlink}
\usepackage{graphicx}

\begin{document}

\title{Long-CLIP: Unlocking the Long-Text \\ Capability of CLIP} 

\titlerunning{Long-CLIP}

\author{Beichen Zhang\inst{\S 1,2}\and
Pan Zhang\inst{1}\and
Xiaoyi Dong\inst{1,3}\thanks{Corresponding author. \ \S \ Work done during an internship in Shanghai AI Laboratory.}\and \\
Yuhang Zang\inst{1}\and
Jiaqi Wang\inst{1}\samethanks
}
\authorrunning{B. Zhang et al.}

\institute{$ ^{1}$Shanghai AI Laboratory  $ ^{2}$Shanghai Jiao Tong University\\
$ ^{3}$The Chinese University of Hong Kong
\\
\email{zhangbeichen@sjtu.edu.cn,\\ \{zhangpan, dongxiaoyi, zangyuhang, wangjiaqi\}@pjlab.org.cn
\\
\url{https://github.com/beichenzbc/Long-CLIP}
}
}


\maketitle
\begin{abstract}
Contrastive Language-Image Pre-training (CLIP) has been the cornerstone for zero-shot classification, text-image retrieval, and text-image generation by aligning image and text modalities. Despite its widespread adoption, a significant limitation of CLIP lies in the inadequate length of text input. The length of the text token is restricted to 77, and an empirical study shows the actual effective length is even less than 20. This prevents CLIP from handling detailed descriptions, limiting its applications for image retrieval and text-to-image generation with extensive prerequisites. To this end, we propose Long-CLIP as a plug-and-play alternative to CLIP that supports long-text input, retains or even surpasses its zero-shot generalizability, and aligns the CLIP latent space, making it readily replace CLIP without any further adaptation in downstream frameworks. Nevertheless, achieving this goal is far from straightforward, as simplistic fine-tuning can result in a significant degradation of CLIP's performance. Moreover, substituting the text encoder with a language model supporting longer contexts necessitates pretraining with vast amounts of data, incurring significant expenses. Accordingly, Long-CLIP introduces an efficient fine-tuning solution on CLIP with two novel strategies designed to maintain the original capabilities, including (1) a \textbf{knowledge-preserved stretching} of positional embedding and (2) a \textbf{primary component matching} of CLIP features. With leveraging just one million extra long text-image pairs, Long-CLIP has shown the superiority to CLIP for about 20\% in long caption text-image retrieval and 6\% in traditional text-image retrieval tasks, \eg, COCO and Flickr30k. Furthermore, Long-CLIP offers enhanced capabilities for generating images from detailed text descriptions by replacing CLIP in a plug-and-play manner. Codes and models are released at \url{https://github.com/beichenzbc/Long-CLIP}.

\end{abstract}

\begin{figure}[ht]
\centering

    \includegraphics[width=1.0\linewidth]{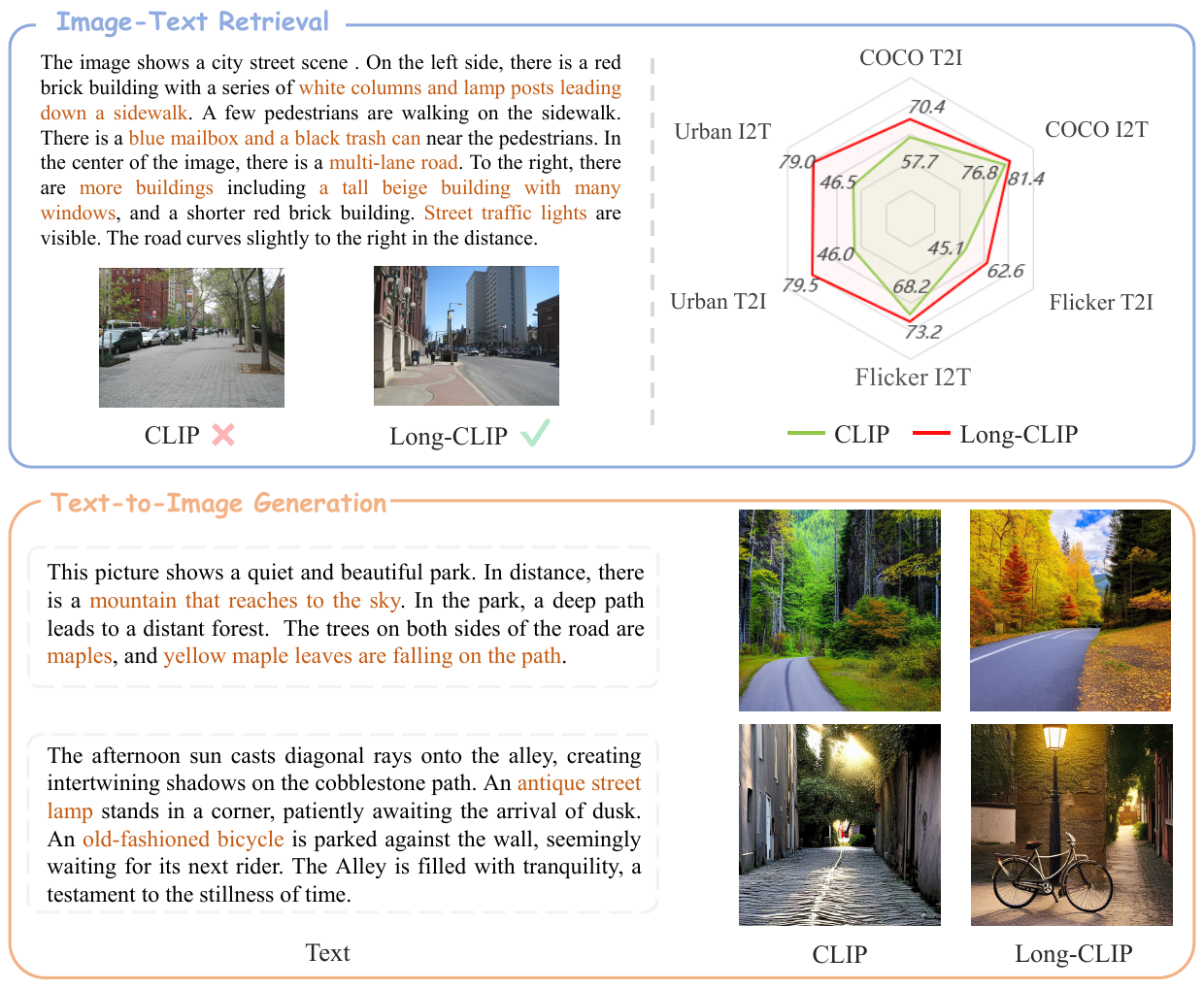}
   \caption{Our Long-CLIP model can capture more detailed attributes and break through the limit of 77 tokens, thus benefiting both \textbf{image-text retrieval} tasks and \textbf{text-to-image generation} tasks. For retrieval, our Long-CLIP model can better capture and represent fine-grained attributes in both modalities, and therefore significantly improves the performance. For image generation, our Long-CLIP model offers enhanced capabilities for generating from detailed text descriptions by replacing the text encoder of CLIP in a plug-and-play manner.} 
   \label{fig:overall}
\end{figure}

\section{Introduction}
\label{sec:intro}

Contrastive Language-Image Pre-training (CLIP)~\cite{clip} is a vision-language foundation model comprising a text encoder and an image encoder. It aligns the vision and language modality based on contrastive learning, widely adopted in downstream tasks, \eg., zero-shot classification~\cite{alphaclip}, text-image retrieval~\cite{lex}, and text-to-image generation~\cite{stablediffusion, clipdraw}.


Nevertheless, the input text length of CLIP is greatly restricted, making it unsuitable for processing detailed descriptions. Specifically, CLIP's text encoder employs an absolute positional embedding limited to 77 tokens, establishing a strict ceiling on input token numbers. Moreover, since the training datasets for CLIP predominantly consist of brief texts, the positional embedding for higher token positions in the CLIP's text encoder are inadequately trained, resulting in an even shorter effective token length. An empirical investigation, detailed in Sec.~\ref{sec:4.1}, reveals that the actual effective length for CLIP is merely 20 tokens.

Furthermore, the absence of a long-text capability not only restricts the potential use of CLIP's text encoder, but also limits the image encoder's ability to capture details and relationships within images. During training, the input text is usually summary text that only contains primary attributes, causing the image encoder to focus on most crucial elements of the image and disregard the other details.
Secondly, CLIP demonstrates limitations in accurately modeling the relationships between diverse attributes. Prior research~\cite{lemon} has highlighted that CLIP utilizes a `bag of concepts' approach for representing various attributes, which can lead to glaring mistakes. For instance, CLIP might assert with high confidence that a lemon is purple when presented with an image containing both a lemon and an eggplant. This significantly undermines its capability to address complex scenarios involving multiple interacting attributes.

Conversely, long texts possess numerous crucial characteristics, evolving fine-grained attributes and indicating the interrelationship between them. Therefore, unlocking the long-text capability of CLIP is of great importance. 


A straightforward strategy to achieve this goal is to release the hard restriction on the input text length via interpolating positional embedding, and then fine-tune CLIP with pairs of images and text data involving long descriptions. However, this strategy involves three crucial limitations: (1) the naive interpolation of positional embedding breaks the well-established representation of short text positions; (2) the image features extracted by the image encoder try to cover all the details to align the long text during fine-tuning, regardless of their varying importance. The overwhelming detailed information in image features after fine-tuning disturbs its alignment with short text; (3) the fine-tuning step will shift the embedding space of CLIP features, leading to extra adaptation cost in downstream frameworks, \eg, Stable Diffusion~\cite{stablediffusion}.  
Experimental results show this simplistic strategy will significantly affect CLIP's short-text capability. The zero-shot classification accuracy on ImageNet will decrease by 13.1\%, and T2I R@1 on COCO will also decrease by 14.4\%. 


To this end, we propose Long-CLIP as a plug-and-play alternative to CLIP. After efficiently fine-tuning CLIP using just an extra 1 million long text-image pairs with only \textbf{0.25} hours on \textbf{8} GPUs, Long-CLIP seamlessly supports long-text input and retains or even surpasses the zero-shot generalizability of CLIP on various benchmarks. Notably, Long-CLIP aligns the CLIP latent space, readily replacing CLIP without any further adaptation in downstream frameworks. To be specific, Long-CLIP introduces two novel designs to achieve these targets, including (1) a \textbf{knowledge-preserved stretching} of positional embedding and (2) a \textbf{primary component matching} of CLIP features. 

In the \textbf{knowledge-preserved stretching} of positional embedding, we first performs empirically study, detailed in Sec.~\ref{sec:4.1}, revealing that the actual effective text length for CLIP is merely 20 tokens. Inspired by this insightful observation, we first keep the first 20 well-trained positional embedding and interpolate the rest 57 insufficiently trained positional embedding by a larger value. This strategy not only improves the overall length but also minimizes the disruption to the well-established position representation. In \textbf{primary component matching} of CLIP features, apart from aligning the fine-grained image feature with a long detailed caption, we also extract the coarse-grained information from the fine-grained image feature and align it with a short summary caption. This requires the model to not only capture different details in an image, but also identify the most important components among them.


Experiments in Sec.~\ref{sec:exp} demonstrate that our performance significantly surpasses the original CLIP. We improve the recall rate by 25\% on long-text image retrieval tasks, and by 6\% on short-text image retrieval tasks. There's no decay on zero-shot classification task. 
Moreover, as we mostly maintain CLIP's latent space, our model can replace the original pre-trained text-encoder in image generating models like Stable Diffusion~\cite{stablediffusion} to unlock its long-text capability in a \textbf{plug-and-play} manner. No additional training is needed.





\section{Related Works}

\textbf{Contrastive Language-Image Pre-training (CLIP).} CLIP~\cite{clip} is a multi-modal model based on contrastive learning. It's training data consists of massive text-image pairs: an image and its corresponding textual description. Through contrastive learning, the model aims to learn the matching relationship between open-world text-image pairs. As CLIP has strong zero-shot generalization abilities, it has been successfully utilized in detection~\cite{vild, glip}, segmentation~\cite{groupvit, lseg}, video understanding~\cite{CLIP4CLIP, videoCLIP} and most commonly, image generating~\cite{DALLE2, clipdraw, vqganCLIP, CLIPpasso}. Many subsequent works have chosen to use the pre-trained vision or text encoder of CLIP. Thanks to its powerful zero-shot generalization ability, these methods can achieve the ability to process open-world information. DALLE-2~\cite{DALLE2}, for example, learns a model for converting CLIP text features to image features, and another model to reconstruct CLIP image features to images, thus achieving open-world text-to-image generation. However, some subsequent works~\cite{rovit, xvlm} have recognized that CLIP lacks the capability of extracting fine-grained information. Therefore, they adopt a similar contrastive method to align the input tokens in a complete sentence with a region of the whole image. However, these works still struggle to capture fine-grained information in a long-caption setting.

\noindent
\textbf{Positional Embedding Interpolation.} To increase the context length, many works~\cite{PI} leverage positional embedding interpolation. Specifically, it linearly down-scales the input position indices to match the original context window size. This is usually a better strategy than extrapolation. However, this is usually applied to relative positional embedding like RoPE~\cite{rope}.

\noindent

\noindent
\textbf{Vision-Language Dataset.} With the development of multimodal capabilities, people are no longer satisfied with fixed-category image datasets like ImageNet~\cite{image} and CIFAR10~\cite{cifar}. Instead, the datasets that contain both images and their corresponding natural language descriptions are required to support open-world applications. The common open-world vision-language datasets include Visual Genome~\cite{VG}, Conceptual-12M~\cite{cc12m}, SBU~\cite{sbu}, COCO~\cite{COCO}, LAION-5B~\cite{laion} and so on. However, these datasets typically only contain short captions. ShareGPT4V~\cite{sharegpt4v} dataset, on the other hand, is a large-scale dataset with 1.2 million long captions. It covers rich information including object properties and spatial relationships. The average character number of the caption reaches 826, which is dozens of times longer than previous datasets.

\section{Method}

\subsection{Exploring the Effective Length of CLIP}
\label{sec:4.1}



In theory, an exceptionally efficient model has the capability to extract all the information from a lengthy caption, regardless of its placement at the beginning, middle, or end of the text. If a model fails to extract information beyond a certain length, we can infer that regardless of the maximum input length it can handle, that specific length represents its true effective limit.

Based on this assumption, we conduct an experiment using urban-200 dataset, a long-caption image-text evaluation dataset constructed by us, which will be elaborated in Sec.~\ref{sec:data}. In theory, the R@1 should also gradually increase as we incrementally increase the input caption length if the model can leverage the additional information. However, our observations reveal that once the input length surpasses 20 tokens, the R@1 of CLIP exhibits slow growth. This suggests that the true effective length of CLIP is no more than 20 tokens, as it struggles to leverage the additional information beyond this point. In contrast, our model's performance continues to improve as the input length increases, reaching its peak at the maximum value within the dataset. This indicates that our model is capable of consistently utilizing the newly added information in the captions.

\begin{figure}[t!]
  \centering
   \includegraphics[width=1.0\linewidth]{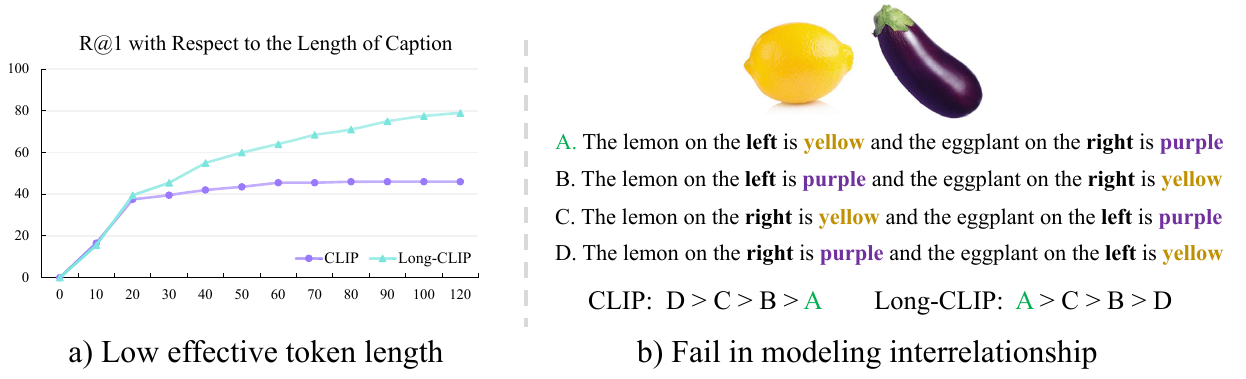}
   \caption{Two major shortcomings of CLIP. The R@1 increases very slowly when the input length exceeds 20 tokens, indicating that the true effective length of CLIP is even no longer than 20 tokens. Moreover, CLIP fails to extract the interrelationship between different attributes, as it assigns the highest similarity score to a representation that mismatches both color and relative position.}
   \label{fig:effective}
\end{figure}

\subsection{Knowledge Preserving Stretching}
Due to the adoption of a learned absolute positional embedding in the text encoder of CLIP, the length of the input text token imposes a rigid constraint.


To address the difficulty of training a new positional embedding, an interpolation strategy is frequently employed. However, this approach often involves a trade-off between supporting longer input lengths and maintaining the integrity of the well-established position representation. Typically, a common strategy is linearly interpolating the positional embedding using a fixed ratio, denoted as $\lambda_{1}$. The calculation for obtaining the new positional embedding $PE^{*}$ is as follows:
\begin{equation}
\begin{split}
 PE^{*}(pos) = (1-\alpha) \times PE(\lfloor \frac{pos}{\lambda_{1}} \rfloor) + \alpha \times PE(\lceil \frac{pos}{\lambda_{1}} \rceil),\quad \alpha = \frac{pos\ \%\ \lambda_{1}}{\lambda_{1}}
 \end{split}
 \label{eq:8}
\end{equation}

where $PE(pos)$ represents the positional embedding for the $pos_{th}$ position and $\alpha$ is a ratio between $0$ and $1$, determining whether the interpolated positional embedding for the $pos_{th}$ position is closer to its preceding or following position.


In this specific task, the straightforward strategy of linear interpolation may not be the most suitable approach. This is because the majority of training texts are likely to be considerably shorter than 77 in the original CLIP model. Therefore the lower positions have been well-trained and can effectively indicate absolute positions, while the higher positions have not received sufficient training and can only provide a rough estimation of relative positions. Consequently, the cost of interpolating the lower positions is much higher compared to interpolating the higher ones, as doing so is more likely to disrupt the well-established representation of absolute position.

Therefore, instead of performing full interpolation with a fixed value, we choose to retain the embedding of the top 20 positions, which aligns with the effective length identified in our experiment. As for the remaining 57 positions, we apply interpolation using a larger ratio denoted as $\lambda_{2}$. This process can be mathematically formulated as follows:

\begin{small}
\begin{equation}
 PE^{*}(pos) = \left\{
 \begin{aligned}
 &PE(pos), \ pos\leq 20\\
 &(1-\alpha) \times PE(\lfloor \frac{pos}{\lambda_{2}} \rfloor) + \alpha \times PE(\lceil \frac{pos}{\lambda_{2}} \rceil),\quad \alpha = \frac{pos\ \% \ \lambda_{2}}{\lambda_{2}}, \ otherwise
 \end{aligned}
 \right .
 \label{eq:9}
\end{equation}
\end{small}
Experiment in Sec.~\ref{sec:4.3} shows that our strategy can improve R@1 on short-text image retrieval by 20\% and even support a longer input length.

\begin{figure}[t!]
  \centering
   \includegraphics[width=\linewidth]{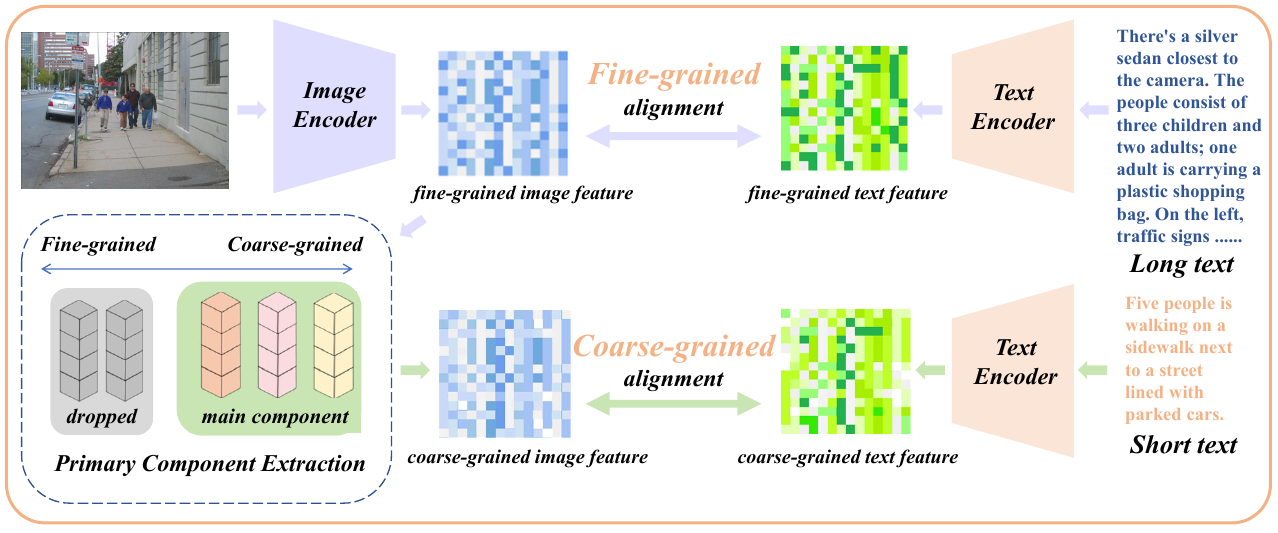}
   \caption{The pipeline of our training process. We align the fine-grained image feature with the a long detailed caption. Moreover, we also apply primary component extraction to keep the main component and extract the coarse-grained image feature. Then, it is aligned with a short summary caption.}
   \label{fig:method}
\end{figure}

\subsection{Fine-tuning with Primary Component matching}

Merely relaxing the strict length constraint on input tokens through positional embedding interpolation is insufficient to fully unlock the long-text capability of the model, as the effective length remains unchanged. Therefore, it is necessary to perform fine-tuning on the CLIP model using long captions.

However, simply fine-tuning the model with long captions will lead to a degradation on the short-text capability. The underlying reason is that a capable model should not only capture various attributes but also understand their relative relationships and differing importance. Merely fine-tuning the model with long captions may push it to another extreme, where it attempts to encompass all attributes in a single image without distinguishing their respective importance.

To this end, we propose a Primary Component matching strategy in long-text fine-tuning to both unlock the long-text capability and maintain the short-text capability. Apart from aligning the fine-grained feature of an image with its long caption, we extract a coarse-grained image feature that focuses on capturing key attributes. This coarse-grained feature is then aligned with a short summary caption. By doing so, we require the model not only to capture detailed attributes but also to discern and prioritize the importance of different attributes.

This strategy enables the model to learn to cover the necessary detailed attributes while also understanding which attributes hold greater significance. By incorporating this approach into the fine-tuning process, we aim to enhance the model's ability to handle both long and short captions effectively.


\begin{figure}[t!]
  \centering
   \includegraphics[width=0.88\linewidth]{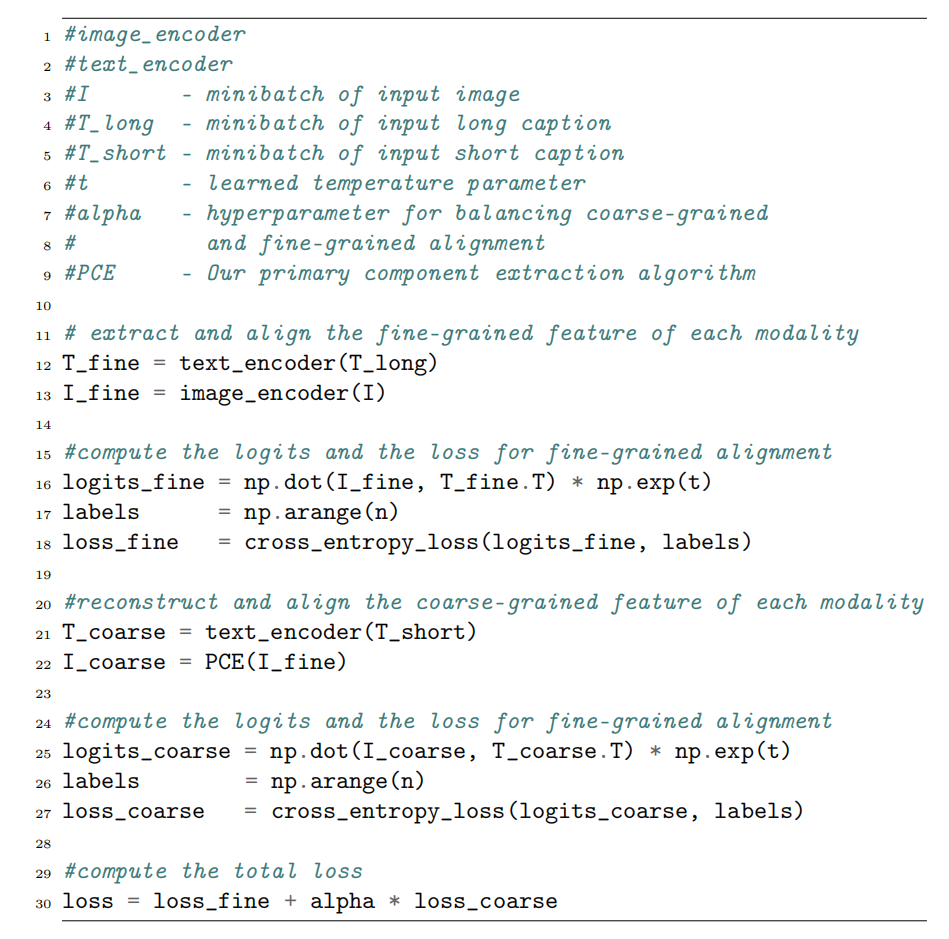}
   \caption{The Numpy-like pseudo-code of our fine-tuning. We separately align the fine-grained and coarse-grained information in both modalities.}
   \label{fig:algorithm}
\end{figure}


Therefore, the key to our task is to find a method for extracting coarse-grained image features from fine-grained image features. This method should be capable of extracting different attributes from the fine-grained image feature and analyzing their importance. To achieve this, we have designed three core modules. The first module is a component-decomposition function $\mathcal{F}$. This function decomposes the feature into several vectors that represent different attributes and also analyzes the importance of each attribute. The second module is a component-filtration function $\mathcal{E}$ which filters out less important attributes based on their analyzed importance. The final module is a component-reconstruction function $\mathcal{F}^{-1}$ which can reconstruct the image feature with different attribute vectors and their corresponding importance.

Given a fine-grained image feature $I_{fine}$, we first extract its different component vectors $v_{t}$ and the corresponding importance $i_{t}$ as Eq.~\ref{eq:decom}

\begin{equation}
 (v_{1}, i_{1}),\ (v_{2}, i_{2}),\ ... ,\ (v_{n}, i_{n}) = \mathcal{F}(I_{fine})
 \label{eq:decom}
\end{equation}

Then, we apply our component-filtration function $\mathcal{E}$ to select the key components and wipe out the others as Eq.~\ref{eq:filter}.
\begin{equation}
 (v_{k_{1}}, i_{k_{1}}),\ (v_{k_{2}}, i_{k_{2}}),\ ... ,\ (v_{k_{m}}, i_{k_{m}}) = \mathcal{E}[(v_{1}, i_{1}),\ (v_{2}, i_{2}),\ ... ,\ (v_{n}, i_{n})],\quad m\ll n
 \label{eq:filter}
\end{equation}

Finally, we apply our component-reconstruction function $\mathcal{F}^{-1}$ to reconstruct the image feature with only the key components and their importance as Eq.~\ref{eq:reconstruct}.

\begin{equation}
 I_{coarse} = \mathcal{F}^{-1}[(v_{k_{1}}, i_{k_{1}}),\ (v_{k_{2}}, i_{k_{2}}),\ ... ,\ (v_{k_{m}}, i_{k_{m}})]
 \label{eq:reconstruct}
\end{equation}

Our total Primary Component Extraction method which extracted the coarse-grained image feature from its fine-grianed one can be represented as Eq.~\ref{eq:total}
\begin{equation}
 I_{coarse} = \mathcal{F}^{-1}(\mathcal{E}(\mathcal{F}(I_{fine}))
 \label{eq:total}
\end{equation}

Thus, the reconstructed image feature only contains the most important coarse-grained information of a given image, and can be aligned with the short summary text properly if the fine-grained image feature can also indicate the true importance of different attributes correctly. Otherwise, if the model can't indicate the true importance, the above process may wipe out the most important attributes while keeping the less important ones, and thus the image feature may fail to match the corresponding short texts.

The implementation of our three core modules can be varied. We utilize a widely-used dimensionality reduction algorithm, Principal Component Analysis method, to achieve the above process. Specifically, the Eigen Value Decomposition (EVD) of the covariance matrix serves as our component-decomposition function $\mathcal{F}$. Then, we select the eigenvectors corresponding to the top 32 largest eigenvalues, which serves as our component-filtration function $\mathcal{E}$. Finally, we reconstruct the image by the linear combination of the selected eigenvectors as component-reconstruction function $\mathcal{F}^{-1}$.

We demonstrate our training pipeline in Fig.~\ref{fig:method} and the pseudo-code in Fig.~\ref{fig:algorithm}.

\section{Experiments}
\label{sec:exp}
\subsection{Experiment Setting}
\textbf{Evaluation Dataset.} \quad We evaluate our model in three downstream tasks.

\textbf{1) zero-shot image classification.} The main dataset is \textit{ImageNet-1K}~\cite{image}. Moreover, \textit{ImageNet-V2}~\cite{imagenetv2}, \textit{ImageNet-O}~\cite{imagenetO}, \textit{CIFAR-10}~\cite{cifar}, \textit{CIFAR-100}~\cite{cifar} are also used to analyze classification ability.

\textbf{2) short-caption image-text retrieval.} 
For traditional short-caption image-text retrieval, we use \textit{COCO2017}~\cite{COCO} and \textit{Flickr30k}~\cite{flickr}. For \textit{COCO2017}, we use the 5k validation set. For \textit{Flickr30k}, to increase difficulty and avoid benchmark saturation, we use the whole 30k dataset instead of the 1k test set. 

\textbf{3) long-caption image-text retrieval.}
For long-caption image-text retrieval, we use random 1k (image, long text) pairs separated from ShareGPT4V~\cite{sharegpt4v}. Moreover, we also collect 200 similar images describing urban scenes and use GPT-4V~\cite{GPT4v} to generate a long caption, which will be elaborated in Sec.~\ref{sec:data}.

\noindent \textbf{Evaluation Setting.} \quad All the experiments in this section share same settings, including the class template in zero-shot classification, all directly truncating the input token if it's longer than restriction, and so on.

\noindent \textbf{Training setting.} \quad We use the ShareGPT4V~\cite{sharegpt4v} dataset as training dataset, which contains about 1M (long caption, image) pairs. The random 1k data is separated as an evaluation dataset. We fine-tune for 1 epoch with batch size 2048. Detailed hyper-parameter settings will be listed in supplementary material.
\begin{figure}[t!]
  \centering
   \includegraphics[width=\linewidth]{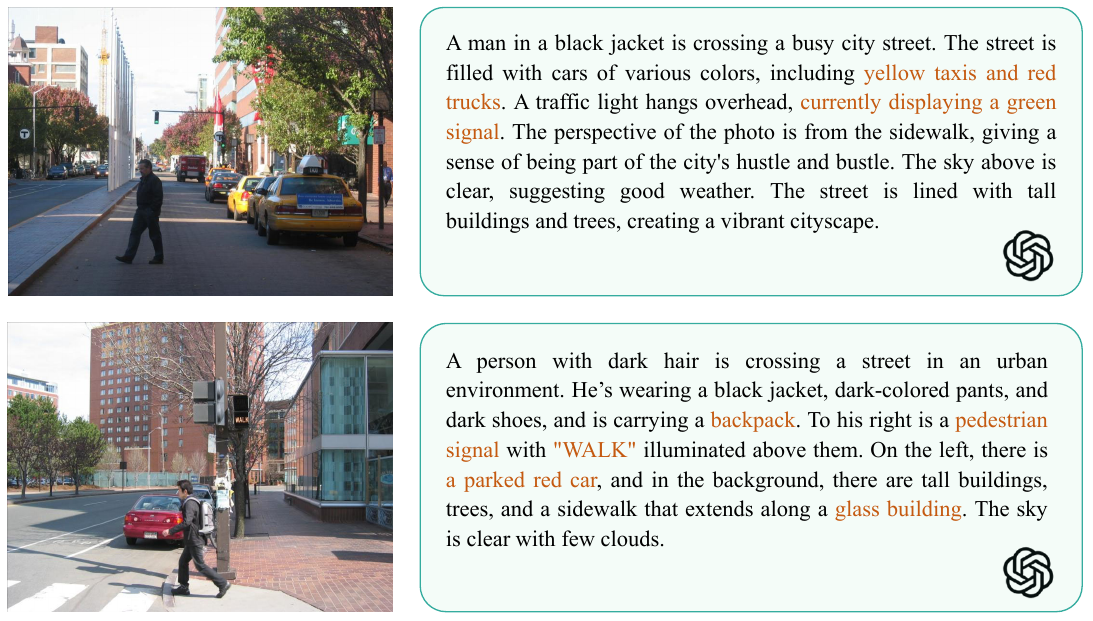}
   \caption{Some (image, long text) pairs of our urban dataset. The images are collected from Visual Genome~\cite{VG} while the long captions are generated by GPT-4V~\cite{GPT4v} and are checked and corrected manually. For these two similar image both showcasing a person crossing the road, the key attributes to distinguish them is marked in brown.}
   \label{fig:dataset}
\end{figure}

\subsection{New Evaluation Dataset on Long Caption Text-Image Retrieval}
\label{sec:data}

Most current image-text retrieval datasets like COCO~\cite{COCO} and Flickr30k~\cite{flickr} only contain short captions and can only evaluate the model on coarse-grained retrieval ability. To better evaluate long-text fine-grained ability, we collect an urban dataset which contains 200 (image, long caption) pairs.

Specifically, we manually select 200 similar images from Visual Genome dataset~\cite{VG}, all showcasing a busy urban view. Then, we use GPT-4V~\cite{GPT4v} to generate a long and complete sentence to describe the image, including types, colors and relative locations of the attributes. The model can successfully match the images with the correct captions only if it successfully understands and models the detailed attributes in both modalities.

The average length of our caption reaches about 101 words, which is far longer than the current existing retrieval datasets.

Fig.~\ref{fig:dataset} showcases some examples of our dataset, which is much harder to distinguish than traditional retrieval datasets, posing a higher demand on the capabilities of the model.

After the first submission, we further scaled up \textbf{Urban-200} into \textbf{Urban-1k}. 
The dataset has been released at \url{https://huggingface.co/datasets/BeichenZhang/Urban1k}. Urban-200 is used in the main paper. Detailed results about Urban-1k is shown in supplementary materials.

\begin{table}[!t]
    \caption{The R@1 of long-caption text-image retrieval on 1k ShareGPT4V~\cite{sharegpt4v} validation set and Urban-200 dataset. Best result is in \textbf{bold}.}
    \centering 
    \resizebox{\textwidth}{!}{
    \begin{tabular}{c| c|  c| c| c|c}
    \toprule
    \multicolumn{2}{c|}{} & \multicolumn{2}{c|}{ShareGPT4V} & \multicolumn{2}{c}{Urban-200} \\
    \multicolumn{2}{c|}{} & ~Image-to-Text~ &  ~Text-to-Image~ &  ~Image-to-Text~ &  ~Text-to-Image~ \\
   
    \midrule
    \multirow{3}{*}{B/16} & CLIP &78.2&79.6&46.5&46.0\\
        & Direct Fine-tuning &94.1&\textbf{93.6}&78.5&78.0\\
        & \textbf{Long-CLIP(Ours)} &\textbf{94.6}&93.3&\textbf{79.5}&\textbf{79.0}\\
    \midrule
    \multirow{3}{*}{L/14} & CLIP & 81.8&84.0&47.0&47.0\\
    & Direct Fine-tuning &95.3&95.4&78.0&76.5\\
        & \textbf{Long-CLIP(Ours)} &\textbf{95.8}&\textbf{95.6}&\textbf{81.5}&\textbf{81.5}\\
    \bottomrule
    \end{tabular}
    }
    \label{tab:longretrieval}
    
    \caption{Results of short-caption text-image retrieval on the 5k COCO2017 validation set and the whole \textbf{30k} Flickr30K dataset. Best result is in \textbf{bold}.}
    \centering 
    \resizebox{\textwidth}{!}{
    \begin{tabular}{c| c|  c  c  c|  c c c| c c c |c c c}
    \toprule
    \multicolumn{2}{c|}{}& \multicolumn{6}{c|}{COCO} & \multicolumn{6}{c}{Flickr30k} \\
    \multicolumn{2}{c|}{}& \multicolumn{3}{c|}{Image-to-Text} &  \multicolumn{3}{c|}{Text-to-Image}&  \multicolumn{3}{c|}{Image-to-Text}&  \multicolumn{3}{c}{Text-to-Image}\\
    \multicolumn{2}{c|}{} & R@1~ & R@5~ &R@10~ & R@1~ & R@5~ &R@10~ & R@1~ & R@5~ &R@10~ & R@1~ & R@5~ &R@10~ \\
    \midrule
    \multirow{3}{*}{B/16} & CLIP &51.8&76.8&84.3&32.7&57.7&68.2&44.1&68.2&77.0    & 24.7&45.1&54.6\\
        & Direct Fine-tuning &37.4&62.3&72.1&21.8&43.4&54.5&25.7&45.8&55.4&17.9&34.5&43.1\\
        & \textbf{Long-CLIP(Ours)} &\textbf{57.6}&\textbf{81.1}&\textbf{87.8}&\textbf{40.4}&\textbf{65.8}&\textbf{75.2}&\textbf{46.8}&\textbf{71.4}&\textbf{79.8}&\textbf{34.1}&\textbf{56.3}&\textbf{65.7}\\
    \midrule
    \multirow{3}{*}{L/14} & CLIP & 56.1&79.5&86.8&35.4&60.1&70.2&48.5&72.6&80.8&28.0&49.3&58.7 \\
    & Direct Fine-tuning &37.9&63.1&72.2&23.1&45.1&55.9&26.0&46.3&55.6&17.9&34.9&43.5\\
        & \textbf{Long-CLIP(Ours)} &\textbf{62.8}&\textbf{85.1}&\textbf{91.2}&\textbf{46.3}&\textbf{70.8}&\textbf{79.8}&\textbf{53.4}&\textbf{77.5}&\textbf{85.3}&\textbf{41.2}&\textbf{64.1}&\textbf{72.6}\\
    \bottomrule
    \end{tabular}
    }
    \label{tab:retrieval}
    
    \caption{Results of zero-shot image classification in the above five validation sets. Best result is in \textbf{bold}.}
    \centering 
    \resizebox{\textwidth}{!}{
    \begin{tabular}{c| c|  c| c| c|c|c|c}
    \toprule
    \multicolumn{2}{c|}{} & ~ImageNet~ & ~ImageNet-O~ & ~ImageNet-V2~ & ~Cifar10~ & ~Cifar100~ & ~Average~ \\
    \midrule
    \multirow{3}{*}{B/16} & CLIP &\textbf{68.4}&42.2&\textbf{61.9}&\textbf{90.8}&67.3&66.12\\
        & Direct Fine-tuning &55.1&31.7&44.8&83.9&59.2&54.94\\
        & \textbf{Long-CLIP(Ours)} &66.8&\textbf{42.7}&61.2&90.7&\textbf{69.3}&\textbf{66.14}\\
    \midrule
    \multirow{3}{*}{L/14} & CLIP & \textbf{75.5}&31.9&\textbf{69.9}&\textbf{95.5}&76.8&\textbf{69.92}\\
    & Direct Fine-tuning &58.4&29.2&52.7&92.7&68.7&60.3\\
        & \textbf{Long-CLIP(Ours)} &73.5&\textbf{33.7}&67.9&95.3&\textbf{78.5}&69.78\\
    \bottomrule
    \end{tabular}
    }
    \label{tab:classification}
\end{table}

\subsection{Comparing with CLIP Model}
\label{sec:4.2}

We compare our model with CLIP in the above three different downstream tasks. The interpolation ratio $\lambda_{2}$ in knowledge-stretching is set 4, extending the maximum input length to 248. To evaluate the scalability of our method, we also fine-tune CLIP ViT-L/14 model using our knowledge-preserved stretching of positional embedding and primary component matching strategy as well. 


Moreover, we also compare with a direct fine-tuning method discussed in Sec.~\ref{sec:intro}, which simply interpolates each position with a fixed ratio $\lambda_{1} = 3$ and only apply (long text, image) pairs in training. 

The detailed result is shown in Tab.~\ref{tab:longretrieval}, Tab.~\ref{tab:retrieval}, and Tab.~\ref{tab:classification}, which shows that our method outperforms our baseline in most aspects. Specifically, for fine-grained task, \ie short or long text-image retrieval, both image and text feature extracted can represent more detailed information of multiple attributes to distinguish from other similar candidates. Therefore, our model can significantly improve performance. For coarse-grained classification tasks, our method doesn't suffer from significant performance degradation like the simple approach thanks to our interpolation and training strategies. This will be discussed in Sec.~\ref{sec:4.3}

\subsection{Ablation Study}
\label{sec:4.3}
\textbf{Effectiveness of our core component.}  
There are two core component strategy in our method. The first is the knowledge-preserved stretching of positional embedding, which prevents a distortion on well-established positional representation. The second is the primary component matching, aligning both coarse-grained and fine-grained information for a text-image pair. To demonstrate that both strategies can improve the performance of the model, we tested the model's accuracy on different tasks both with and without using these two strategies. For the knowledge-preserved stretching (KPS), we compare it with interpolating each position with a fixed ratio 3. For the primary component matching (PCM), we compare it with simply fine-tuning with (long caption, image) pairs. The result is demonstrated in Tab.~\ref{tab:whether}. As can be seen, removing any of the strategies we proposed will result in a significant loss of short text capabilities.

\begin{table}[t]
  \centering
  \caption{A comparison of whether to use our two strategies. Best result is in \textbf{bold}.}
  \resizebox{\textwidth}{!}{
    \begin{tabular}{c|c|c|c|c|c|c}
    \toprule
    ~KPS~ & ~PCM~ & ~ImageNet~ & ~Cifar100~ & ~COCO T2I R@5~ & ~Flickr I2T R@5~ & ~urban T2I R@1~\\
    \midrule
    \XSolidBrush& \XSolidBrush & 55.1& 59.2& 43.4&45.8&78.0\\
    \XSolidBrush& \Checkmark &58.8 & 63.5& 46.1&46.0 &76.5\\
    \Checkmark& \XSolidBrush & 65.6 & 65.9 & 64.3 &70.4 & 78.0\\
    \Checkmark& \Checkmark & \textbf{66.8} & \textbf{69.3} & \textbf{65.8} & \textbf{71.4}&\textbf{79.0}\\
    \bottomrule
    \end{tabular}
    \label{tab:whether}
  }
\end{table}

\begin{table}[t]
  \centering
  \caption{A comparison of different strategies aiming to keep the short-text capability. Best result is in \textbf{bold}.}
  \resizebox{\textwidth}{!}{
    \begin{tabular}{c|c|c|c|c|c}
    \toprule
    Strategy & ~ImageNet~ & ~Cifar100~ & ~COCO T2I R@5~ & ~Flickr I2T R@5~ & ~Urban-200 T2I R@1~\\
    \midrule
    Undistinguished~ & 65.5 & 67.5 & 64.6 & 66.8 & 71.0\\
    Mixed-length text~ & 66.4 & 67.8 & 64.2 & 68.2 & 67.5\\
    Bounded encoding~ & 66.8 & 67.9 & 65.7 & 69.3 & \textbf{80.0}\\
        \textbf{Ours}~ & \textbf{66.8} & \textbf{69.3} & \textbf{65.8} & \textbf{71.4} & 79.0\\
    \bottomrule
    \end{tabular}
  }
\label{tab:short}
\end{table}

\textbf{Different strategy keeping short-text capability.}  
We leverage primary component matching during long-text fine-tuning. And obviously, there exist multiple strategies to help maintain the short-text capability. To further validate the superiority of our method, we design and implement three different strategies.

\textbf{1) Undistinguished Image Feature.} No distinguishing between coarse-grained and fine-grained image features. Align the same image feature $I$ with its corresponding long text $T_{long}$ and short text $T_{short}$ at the same time. 

\textbf{2) Mixed-length Text.}  Randomly replace 10\% long text with the corresponding short text and align the mixed-length text with the image feature.

\textbf{3) Bounded Text Encoder.}   While calculating the long text-image contrastive loss, we also compute the SmoothL1 Loss of the short text features under the current encoder and the frozen original CLIP encoder to ensure that the features of short texts do not have significant deviations.


\begin{figure}[t!]

  \centering
   \includegraphics[width=\linewidth]{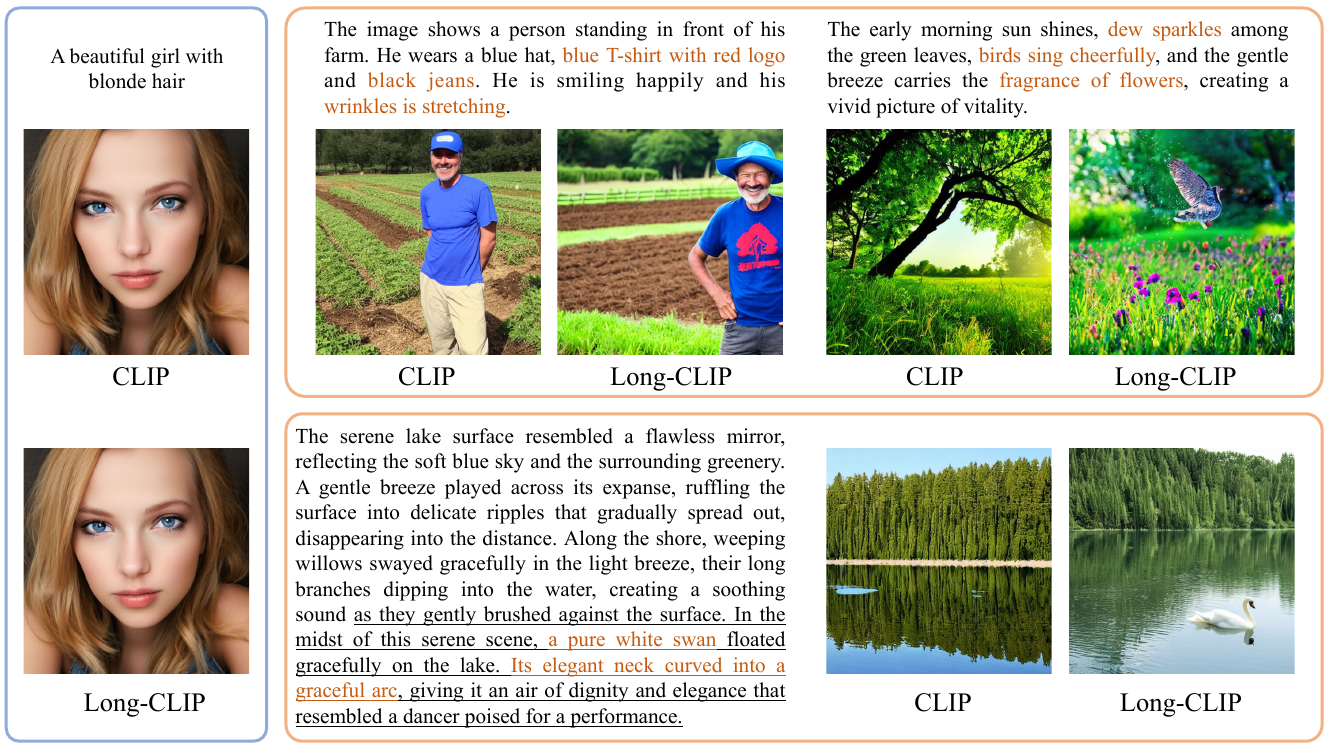}
   \caption{Our Long-CLIP model can benefit the \textbf{text-to-image generation} in a plug-and-play manner in these three aspects. For short simple captions (left), the image generated is nearly the same. For a detailed caption (right top), our model can take in more detailed attributes in it. The caption marked in brown are the detailed attributes missed by the original CLIP, but successfully captured by us. For a long caption (right bottom) exceeding 77 tokens, our model can take in the complete sentence. The underlined caption will be truncated in original CLIP, but can be kept by us.
   }
   \label{fig:demo}
\end{figure}

The detailed result is shown in Tab.~\ref{tab:short}. Ours turns out to outperform others.

Theoretically, the other three strategies also introduce short text in training and can help maintain the short-text capability. The reason why they are less effective is that they still haven't required the model to identify the importance of different attributes. For example, we can observe a significant performance drop on long-text image retrieval tasks for those methods who try to align the same image feature with both long text and short text, \ie Undistinguished and Mixed-length text. The image encoder may be confused as the short text only contains a very limited number of components, which conflicts with our requirements to include all the components in long-text fine-tuning. Ours, instead, requires the model to include all components and identify the most important component. This is consistent with what a picture really implies.

\subsection{Plug-and-Play in Image Generation using CLIP}

CLIP is widely used in many downstream task, including segmentation~\cite{lseg, groupvit}, detection~\cite{glip, vild} and text-to-image generation models like Stable Diffusion~\cite{stablediffusion}. The pre-trained text encoder is often used to extract the text feature of an input to serve as a guidance during image generating. However, just like retrieval tasks, the image generation models using the original CLIP text encoder are still limited by the restriction on 77 tokens and the lack of long-text capability.

Indeed, thanks to the interpolation and matching strategy, we are able to maintain the latent space of the original CLIP model while enhancing its ability to represent fine-grained details in long texts. As a result, our Long-CLIP model exhibits a plug-and-play effect when integrated into existing models. 

For example, we can replace the original CLIP text encoder with our Long-CLIP text encoder in Stable-Diffusion-V1-5~\cite{stablediffusion} with no additional training and keeping other components unchanged. The provided figure (Fig.~\ref{fig:demo}) offers a visual demonstration of the examples, showcasing the enhanced long-text capability of our model. For short and simple text prompts, the images generated by our model closely resemble those generated by the original model. This demonstrates that the short-text capability of the original model is preserved in our approach. However, when presented with longer and more detailed text prompts, our model surpasses the original model by capturing previously overlooked details and including information that exceeds the original 77-token limit.

This plug-and-play nature of our Long-CLIP model allows for seamless integration into various applications and models, provides enhanced capabilities in handling both short and long texts, and supports paragraph-level generation without sacrificing the efficiency and effectiveness of the original CLIP model.

After the first submission, we further use our Long-CLIP model in SDXL~\cite{SDXL}, which will be discussed in detail in supplementary materials.


\section{Conclusion}
We have propose Long-CLIP, a strong and flexible CLIP model with long-text capability. Our model can support text inputs of up to 248 tokens, and can better capture the detailed attributes, obtaining a large improvement on retrieval task. Moreover, our model keeps the performance on zero-shot classification and can replace the CLIP encoder in a plug-and-play manner in image generation task. 

\textbf{Potential Limitations.} Our Long-CLIP model still has an upper bound of input token length, though largely improved. While some relative positional embedding, like RoPE~\cite{rope}, which doesn't pose a hard limit on input token length although the performance may largely decrease for long text input.

\textbf{Scaling-up Potential.} Due to the scarcity of long text-image pairs, we only leverage 1M (image, long-text) pairs in ShareGPT4V~\cite{sharegpt4v}.  This gives us imagination on the scaling-up potential when a large amount of data used. As the sufficient long texts can provide complex information, covering world knowledge, object properties, spatial relationships, and even aesthetic evaluations, the model's ability may be significantly improved.

\section*{Acknowledgments}
This work is partially supported by the National Key R\&D Program of China (2022ZD0160201), and Shanghai Artificial Intelligence Laboratory.



%
%
\bibliographystyle{splncs04}
\bibliography{main}

\clearpage
\title{Supplementary Materials for Long-CLIP: Unlocking the Long-Text Capability of CLIP} 

\titlerunning{Long-CLIP}


\authorrunning{B. Zhang et al.}

    
\author{Beichen Zhang\inst{\S 1,2}\and
Pan Zhang\inst{1}\and
Xiaoyi Dong\inst{1,3}\thanks{Corresponding author. \ \S \ Work done during an internship in Shanghai AI Laboratory.}\and \\
Yuhang Zang\inst{1}\and
Jiaqi Wang\inst{1}\samethanks
}
\authorrunning{B. Zhang et al.}

\institute{$ ^{1}$Shanghai AI Laboratory  $ ^{2}$Shanghai Jiao Tong University\\
$ ^{3}$The Chinese University of Hong Kong
\\
\email{zhangbeichen@sjtu.edu.cn,\\ \{zhangpan, dongxiaoyi, zangyuhang, wangjiaqi\}@pjlab.org.cn
\\
\url{https://github.com/beichenzbc/Long-CLIP}
}
}

\maketitle

\section{Urban-1k Dataset}
\label{sec:urban1k}

Urban-1k is a scaling-up version of Urban-200 dataset in the paper. It contains 1k urban images and their corresponding captions generated by GPT-4V. Each caption contains about 107 words on average. The process for building Urban-1k dataset is exactly the same as Urban-200. It has been released at \url{https://huggingface.co/datasets/BeichenZhang/Urban1k}.

\begin{table}[!h]
  \centering
  \caption{The result on Urban-1k dataset. Best result is in \textbf{bold}.}
  \resizebox{0.7\textwidth}{!}{
    \begin{tabular}{c|c|c|c}
    \toprule
    & Model & ~Urban-1k I2T R@1~ & ~Urban-1k T2I R@1~ \\
    \midrule
    \multirow{2}{*}{B/16~} & CLIP~ & 68.1 & 53.6 \\
    & ~Long-CLIP(Ours)~ & \textbf{78.9} & \textbf{79.5} \\
    \midrule
    \multirow{2}{*}{L/14~} & CLIP~ & 68.7& 52.8 \\
    & ~Long-CLIP(Ours)~ & \textbf{82.7} & \textbf{86.1} \\

    \bottomrule
    \end{tabular}
  }
\label{tab:urban1k}
\end{table}

\section{Insufficient Long-text Ability for CLIP-based Models}
Apart from aligning the whole image with the whole sentence, a few recent works (\eg, PTP-BLIP~\cite{ptpblip} and X-VLM~\cite{xvlm}) have tried to align some text phrases in the whole caption with their corresponding image regions (patches) through contrastive learning and claim to improve the ability to capture fine-grained information.
We evaluate these models on our urban dataset with long captions, and the results are shown in ~\cref{tab:fine}.
We observe that the long-text performance of these methods still has room for improvement since their training datasets predominantly consist of short texts (\eg, Conceptual-12M~\cite{cc12m}, SBU~\cite{sbu}, Visual Genome~\cite{VG} and COCO~\cite{COCO}). By contrast, our Long-CLIP (bottom row) boosts the performance of these baselines by a large margin.

\begin{table}[th]
  \centering
  \caption{The long-text performance of recent advances attempting to improve the fine-grained capability. X-VLM(16M) indicates the version of X-VLM base model with 16M training data.}
  \resizebox{\textwidth}{!}{
    \begin{tabular}{c|c|c|c}
    \toprule
    ~Model~ & ~Image-to-Text~ & ~Text-to-Image~& ~Training Data~\\
    \midrule
    CLIP~\cite{clip} & 46.5& 46.0& WIT\\
    X-VLM(16M)~\cite{xvlm} &53.0&44.5 & CC12M+CC3M+SBU+COCO+VG~\\
    X2-VLM~\cite{x2vlm} & 45.0 & 45.0& CC3M+SBU+COCO+VG+CC12M+LAION~\\
    PTP-BLIP~\cite{ptpblip} & 45.5 & 39.5&CC3M+SBU+COCO+VG~\\
    \textbf{Long-CLIP(Ours)~} & \textbf{81.5} & \textbf{81.5}&WIT+ShareGPT4V\\
    \bottomrule
    \end{tabular}
    \label{tab:fine}
  }
\end{table}

\section{Long-CLIP with SDXL}
~\label{sec:SDXL}
After the first submission, we further use our Long-CLIP model in Stable-Diffusion-XL~\cite{SDXL} in a plug-and-play manner. Specifically, we replace the CLIP-L text encoder with our Long-CLIP-L, and only apply knowledge-preserved stretching (KPS) on Open-CLIP bigG text encoder due to heavy training cost. 

\begin{figure}[!ht]
\centering

    \includegraphics[width=\linewidth]{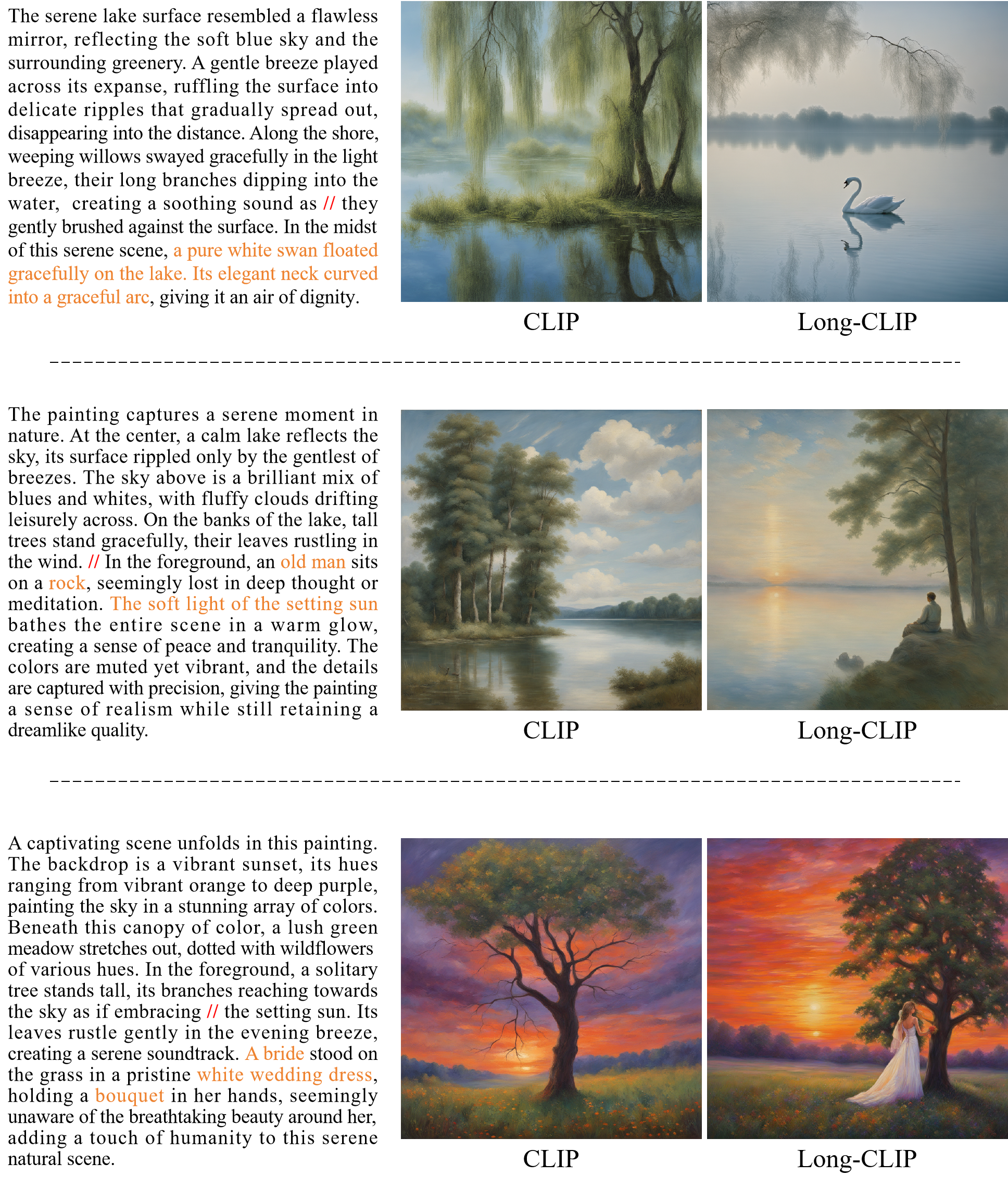}
   \caption{Our Long-CLIP model can help SDXL break through the 77-token limit (marked in \textcolor{red}{red \textbackslash\textbackslash}) and capture the originally truncated attribute (marked in \textcolor{orange}{orange}) with little reduction in the image quality.} 
   \label{fig:SDXL}
\end{figure}

The results are shown in Fig.~\ref{fig:SDXL}, which shows our model can help SDXL break the 77 token limit with little reduction in the image quality.

\section{Generalizability of Proposed Strategy}

We apply our strategy, namely knowledge-preserved stretching and primary component matching, to fine-tune DeCLIP ViT-B/32~\cite{declip}, the result also shows a consistent improvement. 

\begin{table}[h]
\caption{The performance of DeCLIP model and Long-DeCLIP model after fine-tuning. `Avg Cls' means the average accuracy among 5 classification datasets in the main paper. Retrieval tasks are reported in `{T2I}/{I2T}' format with R@5 of short-caption (COCO/Flickr) and R@1 of long-caption (Urban/ShareGPT4V).}
  \centering
  \resizebox{0.75\textwidth}{!}{
    \begin{tabular}{c|c|c|c|c|c}
    \toprule
    Model & ~Avg Cls~ & ~COCO~& ~Flickr~ & ~Urban~ & ~ShareGPT4V~\\
    \midrule
    DeCLIP & 63.6 & ~60.5/45.8~ & ~36.2/25.3~ & ~28.0/25.5~ & ~63.1/60.7~\\
    ~Long-DeCLIP~ & 63.9 & ~61.1/50.7~ & ~36.8/33.8~ & ~54.0/51.0~ & ~84.1/84.3~\\
   
    \bottomrule
    \end{tabular}
  }
\label{tab:short}

\end{table}

\section{Detailed Experiment Setting}
\subsection{Long Text Fine-tuning}
We fine-tune the CLIP model on ShareGPT4V~\cite{sharegpt4v} dataset, which contains about 1M (image, long caption) pairs in total. Detailed hyper-parameter settings in long-text fine-tuning are listed in Tab.~\ref{tab:hyper}. 
\begin{table}[th]
  \centering
  \caption{Detailed hyper-parameters in long-text fine-tuning.}
  \resizebox{0.3\textwidth}{!}{
    \begin{tabular}{c|c}
    \toprule
    ~Hyper-Parameter~ & ~Value~\\
    \midrule
    Batch size & 1024\\
    Training Epochs & 1\\
    Warm-up iterations & 200\\
    Weight decay & 1e-2\\
    Learning Rate & 1e-4\\
    AdamW $\beta_{1}$ & 0.9 \\
    AdamW $\beta_{2}$ & 0.999 \\
    AdamW $\epsilon$ & 1e-8\\
    \bottomrule
    \end{tabular}
    \label{tab:hyper}
  }
\end{table}

\subsection{Zero-shot Classification}
We follow the setup of CLIP~\cite{clip}. For zero-shot classification tasks like ImageNet~\cite{image} and CIFAR-100~\cite{cifar}, we use the 80 pre-defined prompts used in CLIP. We compute the embedding of each class by averaging over the embeddings of the 80 prompts. Then we L2-normalize them. For a given image in a classification dataset, we classify it as the class that has the largest cosine similarity with the image embedding. We use top-1 accuracy as an evaluation metric.

\subsection{Retrieval}
In text-image retrieval tasks, we calculate text-image scores by measuring the cosine similarity between the L2-normalized images and text embedding. We then rank the top-K images for each text caption, as well as the top-K text captions for each image. We use Recall@K as an evaluation metric where K can be 1, 5 and 10, which is a common setting for retrieval tasks.

\section{More Examples on Retrieval and Image Generation}

\begin{figure}[!ht]
\centering

    \includegraphics[width=\linewidth]{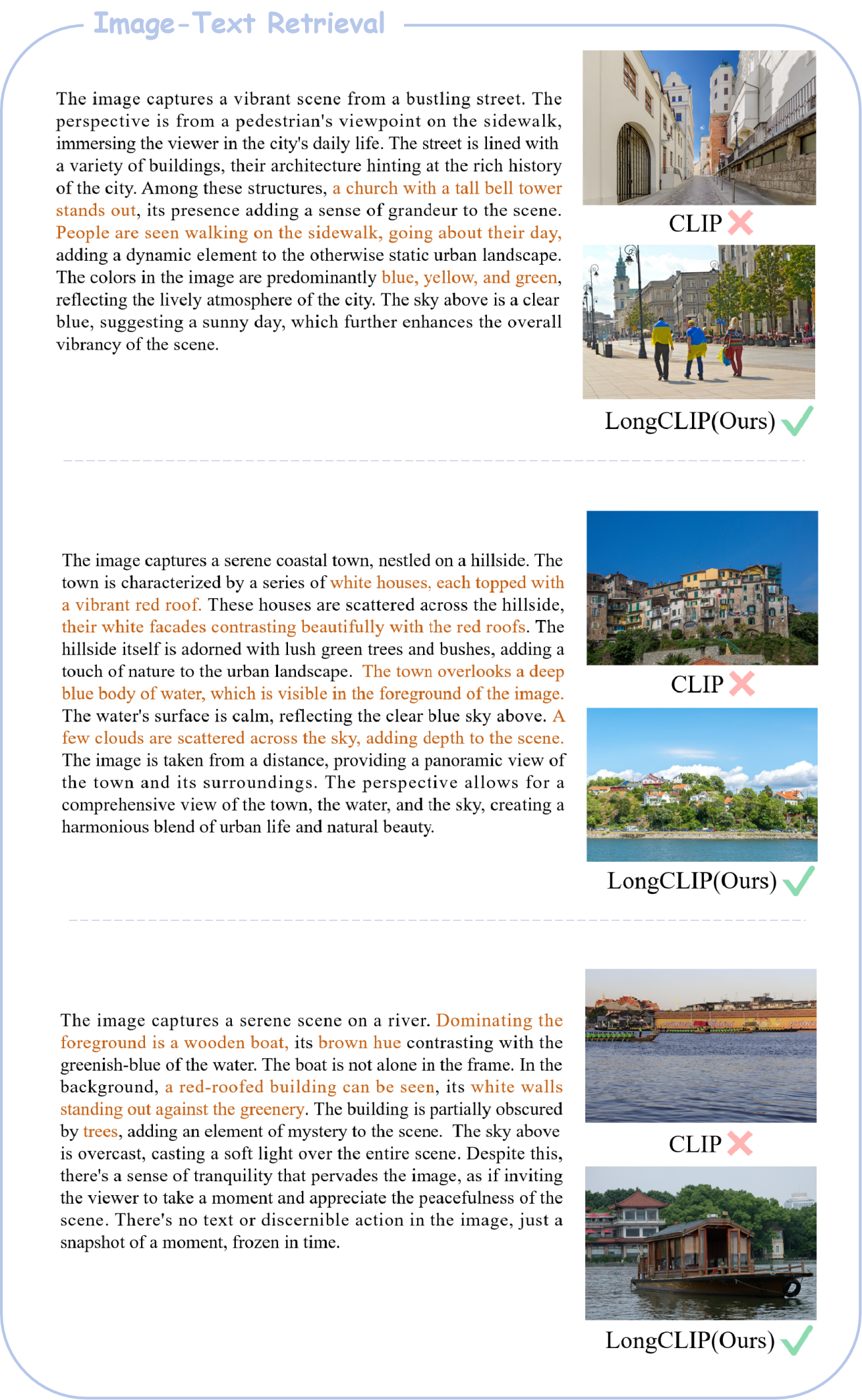}
   \caption{More examples on Image-Text retrieval. The detailed attributes in the long caption to distinguish the correct image is marked in brown.} 
   \label{fig:retrieval}
\end{figure}

\begin{figure}[!ht]
\centering

    \includegraphics[width=\linewidth]{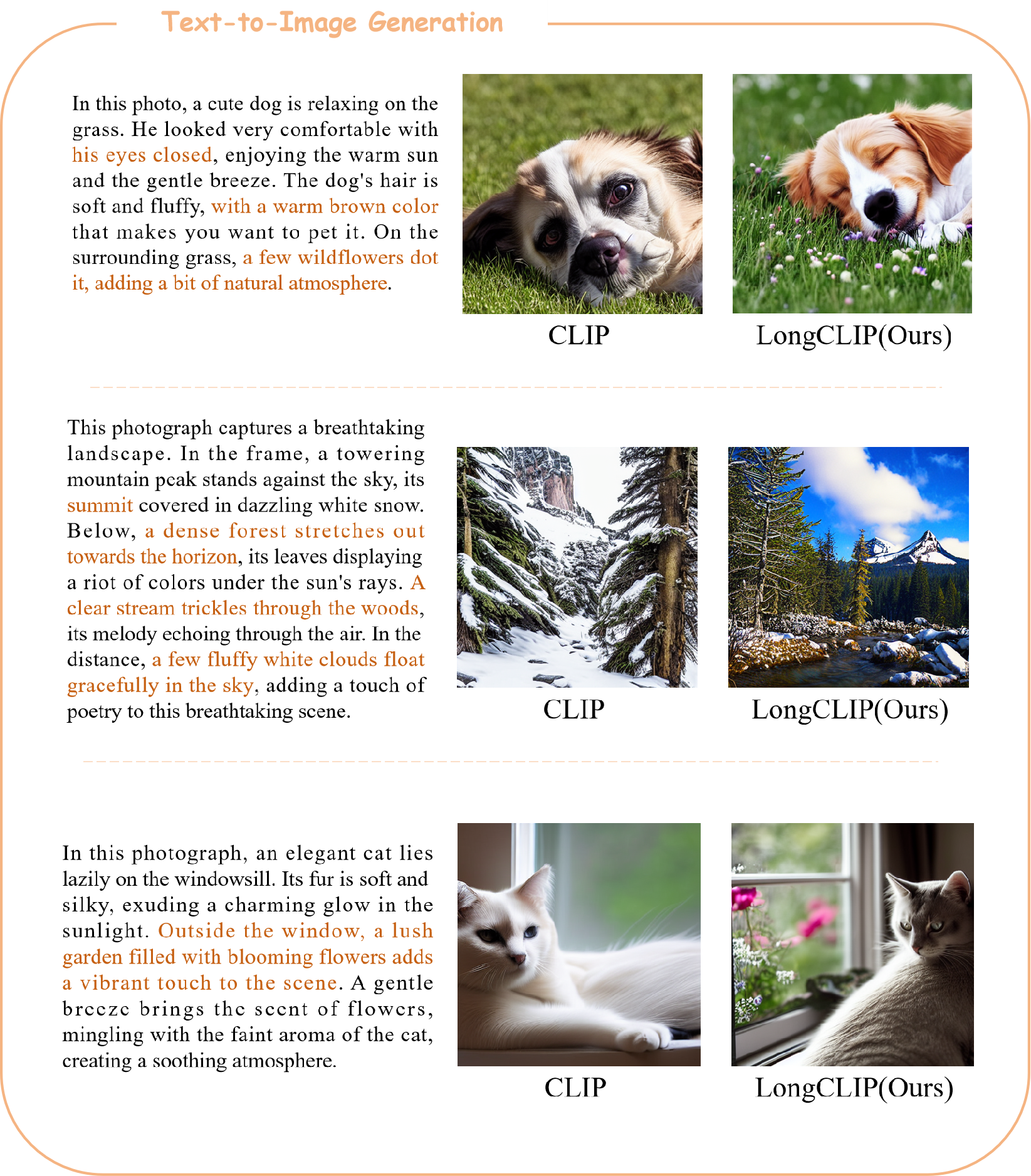}
   \caption{More examples on Text-to-Image Generation. The caption marked in brown are the detailed attributes missed by CLIP, but successfully captured by us.} 
   \label{fig:generation}
\end{figure}

Our Long-CLIP model can capture detailed information in both image and text modalities. Therefore, Long-CLIP can distinguish similar images and texts and improve retrieval accuracy. \cref{fig:retrieval} demonstrates some examples where CLIP fails but our Long-CLIP model can successfully retrieve. Moreover, as shown in \cref{fig:generation}, Long-CLIP can also enhance image generation tasks by covering more details in the text prompt compare to the CLIP baseline.

\end{document}